\pdfoutput=1

\documentclass[11pt]{article}

\usepackage{acl}
\usepackage{graphicx}
\usepackage{times}
\usepackage{latexsym}
\usepackage{subfig}

\usepackage[T1]{fontenc}

\usepackage[utf8]{inputenc}

\usepackage{microtype}

\title{Lexical Squad@Multimodal Hate Speech Event Detection 2023: Multimodal Hate Speech Detection using Fused Ensemble Approach}

\author{Mohammad Kashif, Mohammad Zohair, Saquib Ali \\
        Jamia Millia Islamia, New Delhi, India \\
        \texttt{iamkashif20@gmail.com, mohammadzohair2002@gmail.com}\\
        \texttt{alisaquib95@gmail.com}\\
}

\begin{document}
\maketitle
\begin{abstract}
With a surge in the usage of social media postings to express opinions, emotions, and ideologies, there has been a significant shift towards the calibration of social media as a rapid medium of conveying viewpoints and outlooks over the globe. Concurrently, the emergence of a multitude of conflicts between two entities has given rise to a stream of social media content containing propaganda, hate speech, and inconsiderate views. Thus, the issue of monitoring social media postings is rising swiftly, attracting major attention from those willing to solve such problems. One such problem is Hate Speech detection. To mitigate this problem, we present our novel ensemble learning approach for detecting hate speech, by classifying text-embedded images into two labels, namely \textit{"Hate Speech"} and \textit{"No Hate Speech"}. We have incorporated state-of-art models including InceptionV3, BERT, and XLNet. Our proposed ensemble model yielded promising results with 75.21 and 74.96 as accuracy and F-1 score (respectively). We also present an empirical evaluation of the text-embedded images to elaborate on how well the model was able to predict and classify. We release our codebase here \url{https://github.com/M0hammad-Kashif/MultiModalHateSpeech}
\end{abstract}

\section{Introduction} \label{introduction}

Political events have been a perpetual part of governance to date and serve as a medium of expression for those involved directly or indirectly with the process. But at times, this medium of communication might turn out to be a source of unfortunate insensitive expressions, hate speeches, etc, through verbal forms, visual representations, and physically inconsiderate actions among others. In such cases, it becomes crucial to monitor political events and other potential contributors to the circulation of hate speech and insensitive content.

According to legal publications, hate speech is defined as an expression that seeks to malign an individual for their immutable characteristics, such as their race, ethnicity, national origin, religion, gender, gender identity, sexual orientation, age or disability \citep{carlson2021hate}. Hate speech detection is one of the most important aspects of event identification during political events like invasions \citep{bhandari2023crisishatemm, parihar2021hate}. As is evident in today’s scenario, the incorporation of multimodal data to meet incentives is highly prevalent and is a major concern for hate speech detection and analysis.

In this paper, we elaborate on our submission for the shared task\footnote{\url{https://emw.ku.edu.tr/case-2023/}}, for multimodal hate speech detection through text-embedded images from the Russia-Ukraine war, which is a part of the bigger picture leading to a significantly demanding issue \cite{thapa2023multimodal}. Multimodal content being advertised through physical spaces, social media, etc, is a mode of spreading hate speech and spiteful views being used extensively in the current scenario. A significant contributor to this phenomenon is the sharing of text-embedded images, representing the views of an individual or a group of individuals, either directly or indirectly. In accordance with this fact, we aim to categorically determine if a given text-embedded image conveys hate speech in any possible form or not.

The rest of the paper is structured as follows: Section \ref{literatureReview} illustrates the existing work which has been carried out in this field of research; Section \ref{dataset&task} describes the dataset and task for our research study; Section \ref{modelArchitecture} elaborates our proposed model architecture including the individual blocks incorporated in the same; Section \ref{results} states the results obtained from this work along with its empirical analysis; Section \ref{conclusion} provides a view of the future scope in this domain besides concluding the paper.

\section{Literature Review} \label{literatureReview}

Extensive work has been carried out to survey the extent of incorporating technology for hate speech detection. For instance, in \citep{schmidt2017survey}, a survey has been carried out on the scope of hate speech detection using natural language processing. Through this study, the features, terminologies, existing approaches, and techniques in this context have been highlighted.

Another similar research work \citep{abro2020automatic}, shows the comparison of the performance of three feature engineering techniques and eight machine learning algorithms on a publicly available dataset having three distinct classes. The results of this research work showed that the bigram features when used with the support vector machine algorithm best performed with 79\% off overall accuracy.

In another study \citep{badjatiya2017deep}, an experiment has been performed to emphasize the usage of deep learning for hate speech detection in tweets. A Twitter dataset containing relevant tweets has been used to classify them as being racist, sexist, or neither. As per the results obtained in this study, the deep learning methods outperform state-of-the-art char/word n-gram methods by $\sim$18 F1 points.

In recent times, multiple attempts have been made to deal with the concern of intelligently determining the spread of hate speech and related expressions through multimodal data. For instance, as a part of the research study \citep{gomez2020exploring}, it was attempted to jointly analyze textual and visual information for hate speech detection, using a large-scale dataset from Twitter, MMHS150K. The researchers associated with this study have compared the implementation of models working on multimodal data with those on unimodal data.

Another research work \citep{das2020detecting}, featuring the detection of hate speech in multimodal memes, forms its basis for categorizing a meme as hateful or non-hateful. As a part of this, the visual modality using object detection and image captioning models to fetch the “actual caption” has been explored and combined with the multi-modal representation to perform binary classification. Along with this, an effort has been made to enhance the predictions using sentiment analysis.

Another instance of research work \citep{velioglu2020detecting}, has been carried out on a dataset containing more than 10000 examples of multimodal content, wherein VisualBERT, which is meant to be the “BERT of vision and language” was trained multimodally on images and captions and was augmented with Ensemble Learning.

\section{Dataset and Task} \label{dataset&task}

As a part of The 6th Workshop on Challenges and Applications of Automated Extraction of Socio-political Events from Text (CASE @ RANLP 2023), the Sub-task A for this research experimentation is to identify whether the given text-embedded image contains hate speech or not \cite{thapa2023multimodal}.

The dataset \cite{thapa-etal-2022-multi} provided for this task consists of around 4700 text-embedded images, having annotations for the prevalence of hate speech. As a two-way classification task, the two classes in the given dataset correspond to “Hate Speech” and “No Hate Speech”, with 2665 and 2058 samples corresponding to the respective classes. The training data consists of 1942 and 1658 samples against the "Hate Speech" and "No Hate Speech" labels (respectively). Concurrently, the evaluation and testing data consists of 443 random samples each. All the images have a unique identifier called "index".

In the training data, the classes are well-balanced, implying the occurrence of 50 text-embedded images each against the two labels, that is, “Hate Speech” as well as “No Hate Speech”.

\section{Model Architecture} \label{modelArchitecture}
This section describes the proposed model architecture for classifying text-embedded images as "Hate Speech" or "No Hate Speech".

\begin{figure*}[htbp]
    \centering
    \includegraphics[width=\textwidth]{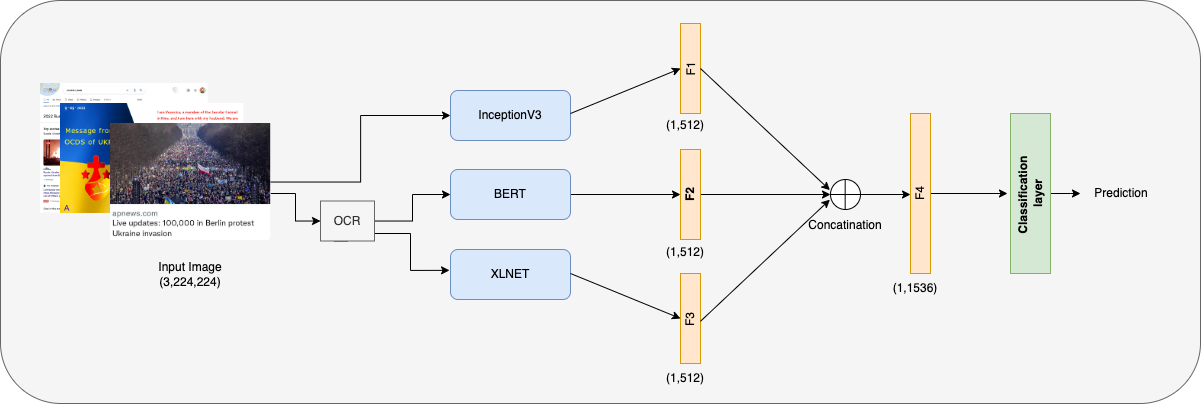}
    \caption{Proposed ensemble architecture }
    \label{fig:ensembleModel}
\end{figure*}

As depicted in Figure \ref{fig:ensembleModel}, we hereby propose an ensemble approach for this binary classification problem. Due to the multimodal nature of the data, it is necessary to extract both visual and textual features from the provided content containing the same. To comprehend the context of a text-embedded image, it is necessary to map the textual context to its visual context. So as to have both of these contexts, we propose an ensemble model that extracts both of these characteristics from an image.

We have incorporated respective models based on convolutional neural networks (CNN), and pre-trained transformer models, which provided good results on the given dataset. InceptionV3 optimizes the neural network for better adaptation as it has a deeper network compared to its predecessors and uses auxiliary classifiers as regularizers. Eventually, it serves as an enhanced option for visual comprehension required for this task. Since transformer models are based on the mechanism of self-attention and differentially weigh the significance of each part of the input textual data, they serve as an ideal option for the textual comprehension of this task.

Our model is comprised of three backbones, one of which (InceptionV3) extracts visual information while the other two (BERT and XLNet) extract textual information.

\subsection{InceptionV3}
Inception-v3 is a convolutional neural network architecture from the Inception family that makes several improvements including using Label Smoothing, Factorized 7 x 7 convolutions, and the use of an auxiliary classifier to propagate label information lower down the network (along with the use of batch normalization for layers in the sidehead) \cite{szegedy2016rethinking}.

The InceptionV3 architecture uses a novel "Inception module" that extracts multi-scale features using various-sized convolutional filters in the same layer \cite{szegedy2015going}. In order to improve the learning of representations, this module enables the network to capture both local and global contextual information.

Apart from that, the Inception module employs 1x1 convolutions along with dimensionality reduction strategies to lessen computational complexity. The inception block takes the input image \[ I \in  R ^ {C*H*W} \] and outputs a feature vector of
\[ F1 \in R ^ {1*512}\]

\subsection{BERT}
BERT’s model architecture is a multi-layer bidirectional Transformer encoder, designed to create state-of-the-art models for a wide range of tasks, such as question answering and language inference, without substantial task-specific architecture modifications \cite{devlin2018bert}.

We have incorporated BERT into our proposed ensemble model to extract textual features. As a result of being trained on a large corpus of unlabeled text, BERT has a solid language foundation and a better understanding of general language representation. We have extracted text from the image using Google's Tesseract-OCR Engine \cite{smith2007overview}, which is eventually tokenized and fed to the BERT model.

BERT outputs a feature vector of size 1x768 which is then provided to the linear layer to generate feature vector  \[ F2 \in R ^ {1*512} \]

\subsection{XLNet}
XLNet is a pre-trained transformer model, which includes segments recurrence, introduced in Transformer-XL \cite{yang2019xlnet, dai2019transformer}, allowing it to digest longer documents \cite{shaheen2020large}.

In order to pay greater attention to text features, we have incorporated a language model into our ensemble-learning model, yet again. XLNet surpasses the limitations of conventional autoregressive models by taking into account all possible permutations of words in a sentence, resulting in enhanced language representation and comprehension.

XLNet is based on the pretraining and fine-tuning paradigm and utilizes the Transformer architecture. The extracted text, from the OCR Engine \cite{smith2007overview}, is fed to the tokenizer and XLNet, which generate a 1x768-dimensional feature vector, which is then fed to the linear layer, which generates a
\[ F3 \in R ^ {1*512} \]

\subsection{Ensemble Model}
Ensemble learning or ensemble model is the combination of numerous different predictions from different models to make the final prediction \cite{ganaie2022ensemble}. This has always been an elegant way of enhancing the performance of models.

Stacking is one of the ensemble learning integration approaches in which the meta-learning model is utilized to integrate the output of base models \cite{dvzeroski2004combining, zohair2022innovators}. Following this strategy, we incorporated our implemented individual models into the blueprint of a stacked ensemble model. This required the generation of individual embeddings from respective models as described in the preceding subsections.

The embeddings F1, F2, and F3 as obtained from the InceptionV3, BERT, and XLNet (respectively), are concatenated to form F4 as the final embedding for the meta-layer \cite{sesmero2015generating}.
\[ F4 \in R ^ {1*1536} \]  

After the final embedding (F4), has been processed through the linear layer, a feature vector of size 128 is eventually produced. This feature vector is then forwarded to the final linear layer for classification, which eventually enhances the accuracy of the predictions. After every linear layer, a ReLU non-linearity is applied \cite{agarap2018deep}.

\subsection{Hyperparameter}
Some of the hyperparameters were kept constant in all models, namely a learning rate of 3e-4, regularization factor of 3e-5, vocab size of 512, and StepLR as the learning rate scheduler.

For our training, we utilized the Adam optimizer \cite{zhang2018improved} and trained the model for 100 epochs. All experiments were conducted on a system equipped with an NVIDIA-A100 GPU, augmented by 64 GB of RAM, with Ubuntu 20.04 as the operating system. The implementation was carried out using the PyTorch framework.

\subsection{Loss and Metric Used}

We have used the weighted cross-entropy loss to penalize the ensemble model with more effectiveness during its training for multimodal classification \cite{phan2020resolving}.  In addition, we have used accuracy as a standard performance metric for comparing models.

\section{Results and Discussion} \label{results}
We have mentioned the results obtained on the validation data in Table \ref{table:tab1}. The results corresponding to the submission were obtained on the test data provided for this sub-task, which have been reflected in Table \ref{resultsTable}. As it is evident from the quoted metrics in Table \ref{table:tab1}, achieved after careful experimentation for the desired task, the proposed ensemble model outperformed various individual models which have been brought into usage for classifying hate speech in the provided dataset.

\begin{table}[htbp]
\centering
\begin{tabular}{|c|c|c|c|} 
\hline
\textbf{Model}     & \textbf{Accuracy}   & \textbf{F1 Score}    \\ 
\hline
BERT                & 69.65        & 69.51            \\ 
\hline
XLNET          & 71.80          & 71.56            \\ 
\hline
InceptionV3         & 48.12          & 48.11            \\ 
\hline
MobileNetV3          & 42.41          & 42.20            \\ 
\hline
ResNet 152             & 44.47          & 44.38           \\ 
\hline
BERT + XLNET & 73.51 & 73.39 \\
\hline
\textbf{Ensemble Model} & \textbf{75.21} & \textbf{74.96}   \\
\hline
\end{tabular}
\caption{Metric Comparison for proposed ensemble model with conventional models}
\label{table:tab1}
\end{table}

\begin{table*}[t]
    \centering
    \begin{tabular}{|c|c|c|c|c|c|c|}
    \hline
    \textbf{\#} & \textbf{User} & \textbf{<Rank>} & \textbf{Recall} & \textbf{Precision} & \textbf{F1} & \textbf{Accuracy} \\
    \hline
    1 & arc-nlp & 1.0000 & 0.8567 (1) & 0.8563 (1) & 0.8565 (1) & 0.8578 (1) \\ \hline
    2 & bayesiano98 & 2.0000 & 0.8561 (2) & 0.8528 (2) & 0.8528 (2) & 0.8533 (2) \\ \hline
    3 & karanpreet\textunderscore singh & 3.0000 & 0.8508 (3) & 0.8476 (3) & 0.8463 (3) & 0.8465 (3) \\ \hline
    4 & DeepBlueAI & 4.0000 & 0.8356 (4) & 0.8335 (4) & 0.8342 (4) & 0.8352 (4) \\ \hline
    5 & csecudsg & 5.0000 & 0.8252 (5) & 0.8244 (5) & 0.8248 (5) & 0.8262 (5) \\ \hline
    6 & Jesus\textunderscore Armenta & 6.0000 & 0.8121 (6) & 0.8094 (6) & 0.8097 (6) & 0.8104 (6) \\ \hline
    7 & Avanthika & 7.0000 & 0.7878 (7) & 0.7881 (7) & 0.7880 (7) & 0.7901 (7) \\ \hline
    8 & Sarika22 & 8.0000 & 0.7806 (8) & 0.7849 (8) & 0.7821 (8) & 0.7856 (8) \\ \hline
    9 & rabindra.nath & 9.0000 & 0.7768 (9) & 0.7842 (9) & 0.7788 (9) & 0.7833 (9) \\ \hline
    10 & md\textunderscore kashif\textunderscore 20 & 10.0000 & 0.7270 (10) & 0.7372 (10) & 0.7287 (10) & 0.7359 (10) \\ \hline
    11 & lueluelue & 11.7500 & 0.5219 (12) & 0.5219 (12) & 0.5219 (11) & 0.5260 (12) \\ \hline
    12 & pakapro & 12.7500 & 0.4938 (13) & 0.4939 (13) & 0.4936 (12) & 0.4966 (13) \\ \hline
    13 & Sathvika.V.S & 11.5000 & 0.5334 (11) & 0.7240 (11) & 0.4294 (13) & 0.5779 (11) \\ \hline
    \end{tabular}
    \caption{Rank Table (Sub-Task A)}
    \label{resultsTable}
\end{table*}

\subsection{Metric Comparison}
The text-based models, including BERT and XLNet, gave an accuracy of 69.65 and 71.80 (respectively) when implemented individually with respect to the given dataset. On the other hand, the image-based models including InceptionV3, MobileNetV3, and ResNet 152, gave accuracy levels of 48.12, 42.41, and 44.47 (respectively) for the same set of data. The combination of BERT and XLNet (without the visual component) gave an accuracy of 73.51.

With regard to this, our proposed ensemble model, developed with InceptionV3, BERT, and XLNet as its individual blocks, provided promising results with an accuracy of 75.21 and an F-1 score of 74.96 on the given dataset, as quoted in Table \ref{table:tab1}.

Our model outperforms the existing works oriented towards multimodal hate speech detection, with an overall accuracy of 75.21. For instance, in \cite{das2020detecting}, the proposed system achieved the best accuracy of 68.4. In \cite{velioglu2020detecting}, the proposed model, VisualBERT achieved an accuracy of 70.93.

The baseline accuracy and F-1 score for the given sub-task are 79.8 and 78.6 (respectively). The proposed model's performance metrics are comparable to the median accuracy and F-1 score of 79.01 and 78.8 (respectively). The same has been mentioned in Table \ref{baseline}.

\vspace{1em}
\begin{table}[htbp]
\centering
\begin{tabular}{|c|c|c|} 
\hline
\textbf{Model}     & \textbf{Accuracy}   & \textbf{F1 Score}    \\ 
\hline
Baseline & 79.8 & 78.6 \\
\hline
Median & 79.01 & 78.8 \\
\hline
\textbf{Proposed} & 75.21 & 74.96 \\
\hline
\end{tabular}
\caption{Metric Comparison for proposed ensemble model with median and baseline scores}
\label{baseline}
\end{table}

\begin{figure}[htbp]
    \centering
    \includegraphics[width=\columnwidth]{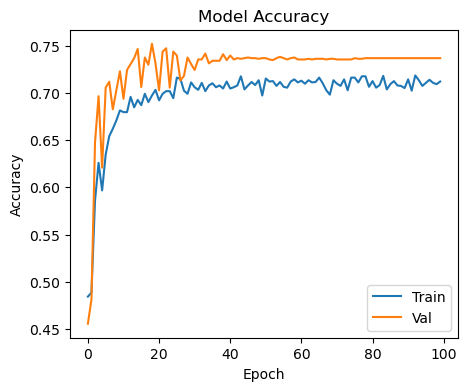}
    \caption{Variation of accuracy with respect to epochs for the \textbf{proposed ensemble model}}
    \label{fig:boat1}
\end{figure}

\begin{figure}[htbp]
    \centering
    \includegraphics[width=\columnwidth]{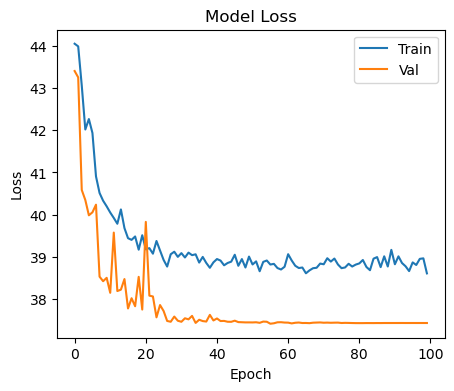}
    \caption{Variation of loss function with respect to epochs for the \textbf{proposed ensemble model}}
    \label{fig:boat2}
\end{figure}

The variation of accuracy level with respect to the number of epochs taken for model training has been depicted in Figure \ref{fig:boat1}. Along with this, the variation of the loss function with respect to the number of epochs taken for model training has been depicted in Figure \ref{fig:boat2}.

\subsection{Empirical Analysis}
In this section, we provide an empirical analysis of our model's predictions for the sample instances of text-embedded images to elaborate on the precision yielded by the model as per the desired task.

\begin{figure}[htbp]
\centering
\begin{tabular}{cccc}
\subfloat[]{\includegraphics[width = 1.4in,height=1.4in]{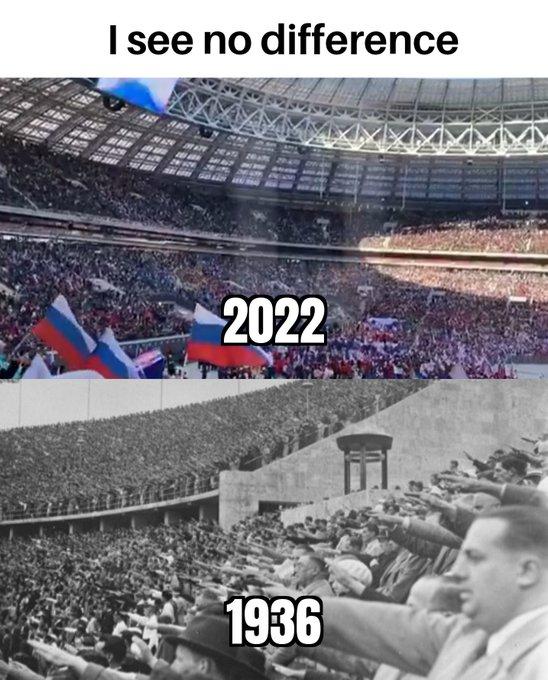} \label{4(a)}} &
\subfloat[]{\includegraphics[width = 1.4in,height=1.4in]{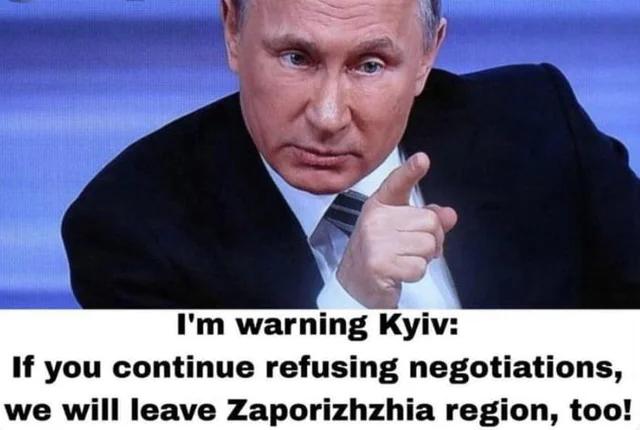} \label{4(b)}} & \\
\subfloat[]{\includegraphics[width = 1.4in,height=1.4in]{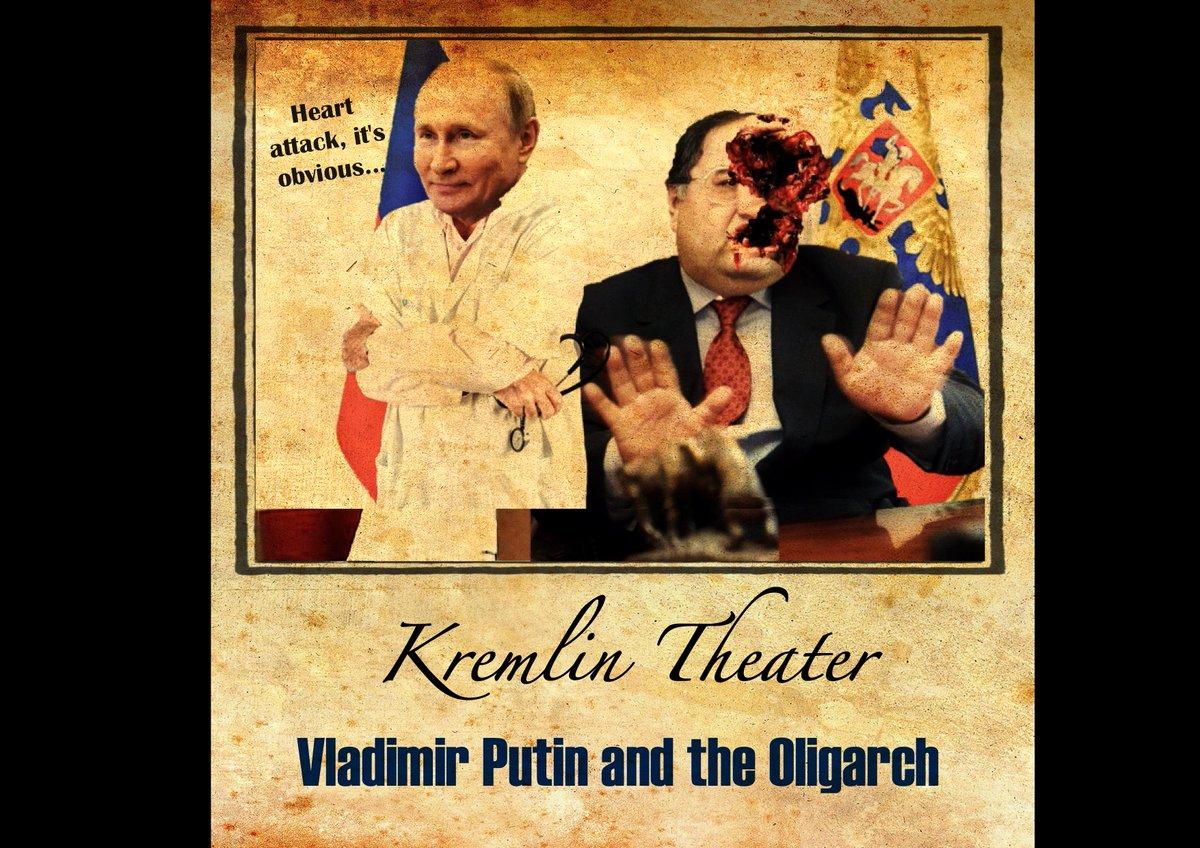} \label{4(c)}} &
\subfloat[]{\includegraphics[width = 1.4in,height=1.4in]{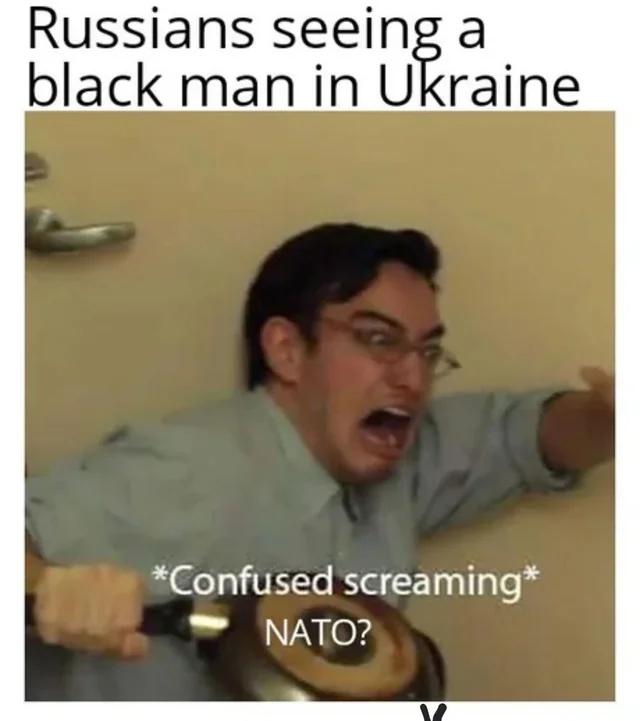} \label{4(d)}}
\end{tabular}
\caption{Sample hate speech images}
\label{hatespeech}
\end{figure}

\begin{figure}[htbp]
\centering
\begin{tabular}{cccc}
\subfloat[]{\includegraphics[width = 1.4in,height=1.4in]{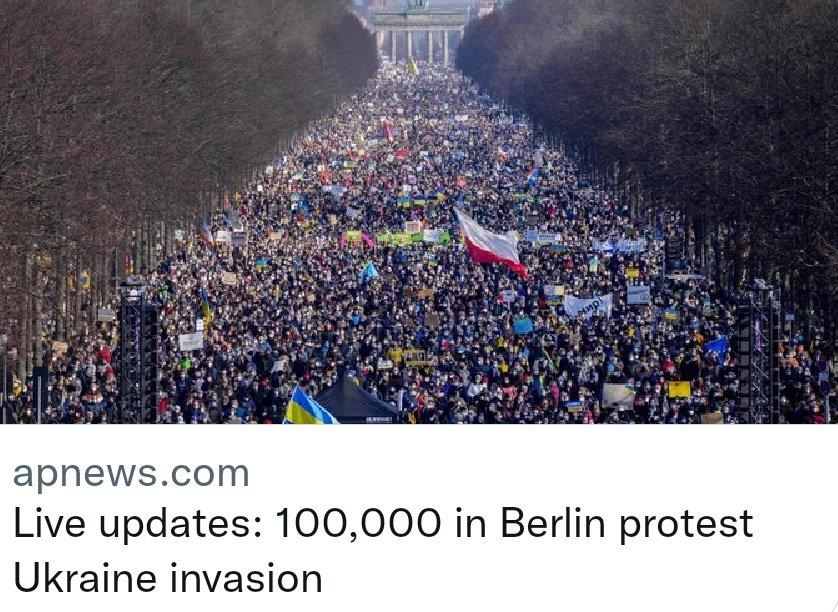} \label{5a}} &
\subfloat[]{\includegraphics[width = 1.4in,height=1.4in]{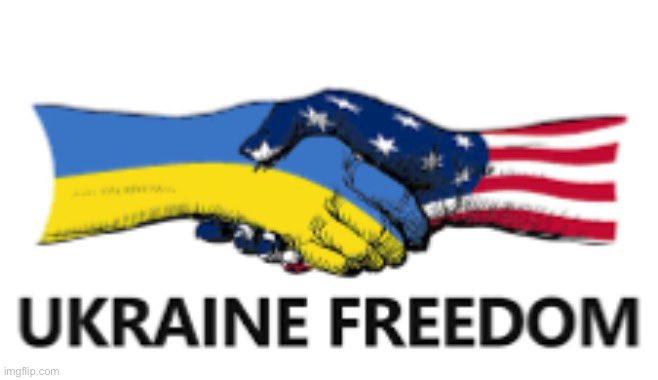} \label{5b}} & \\
\subfloat[]{\includegraphics[width = 1.4in,height=1.4in]{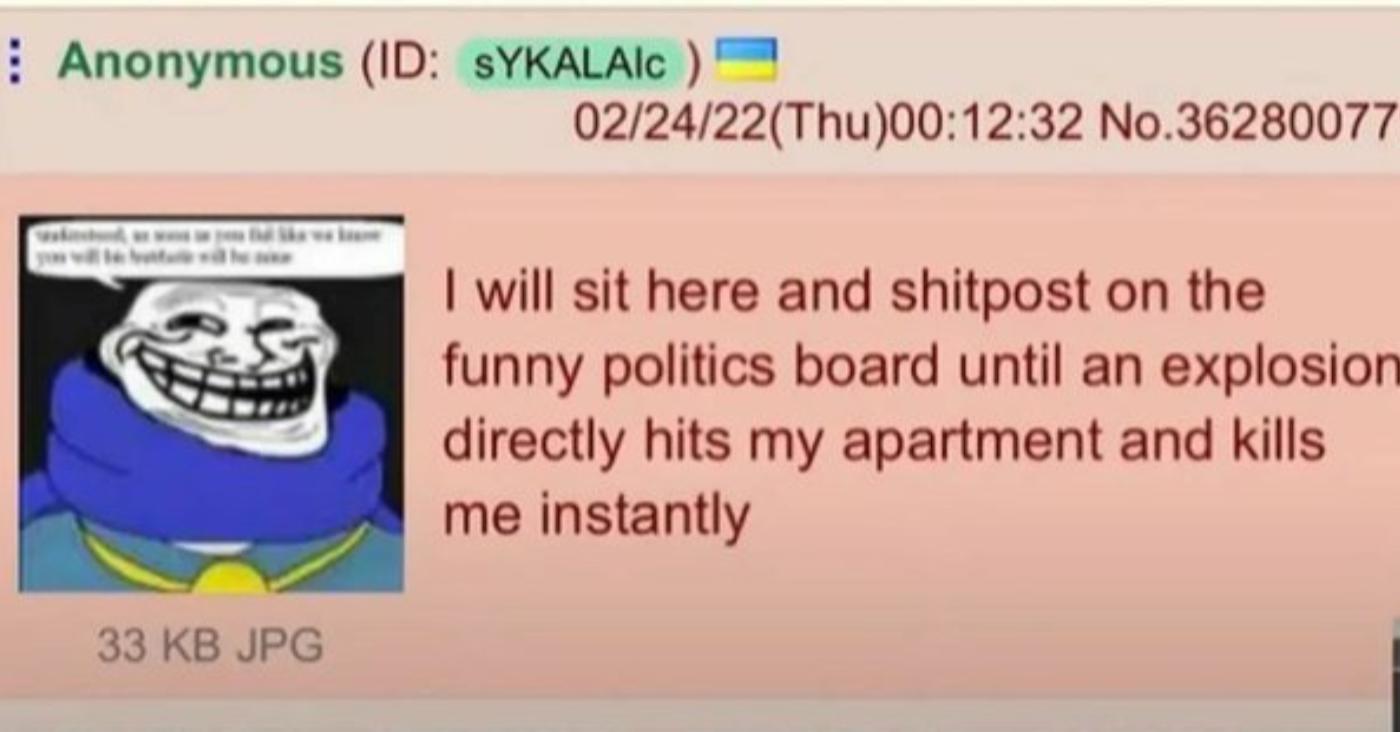} \label{5c}} &
\subfloat[]{\includegraphics[width = 1.4in,height=1.4in]{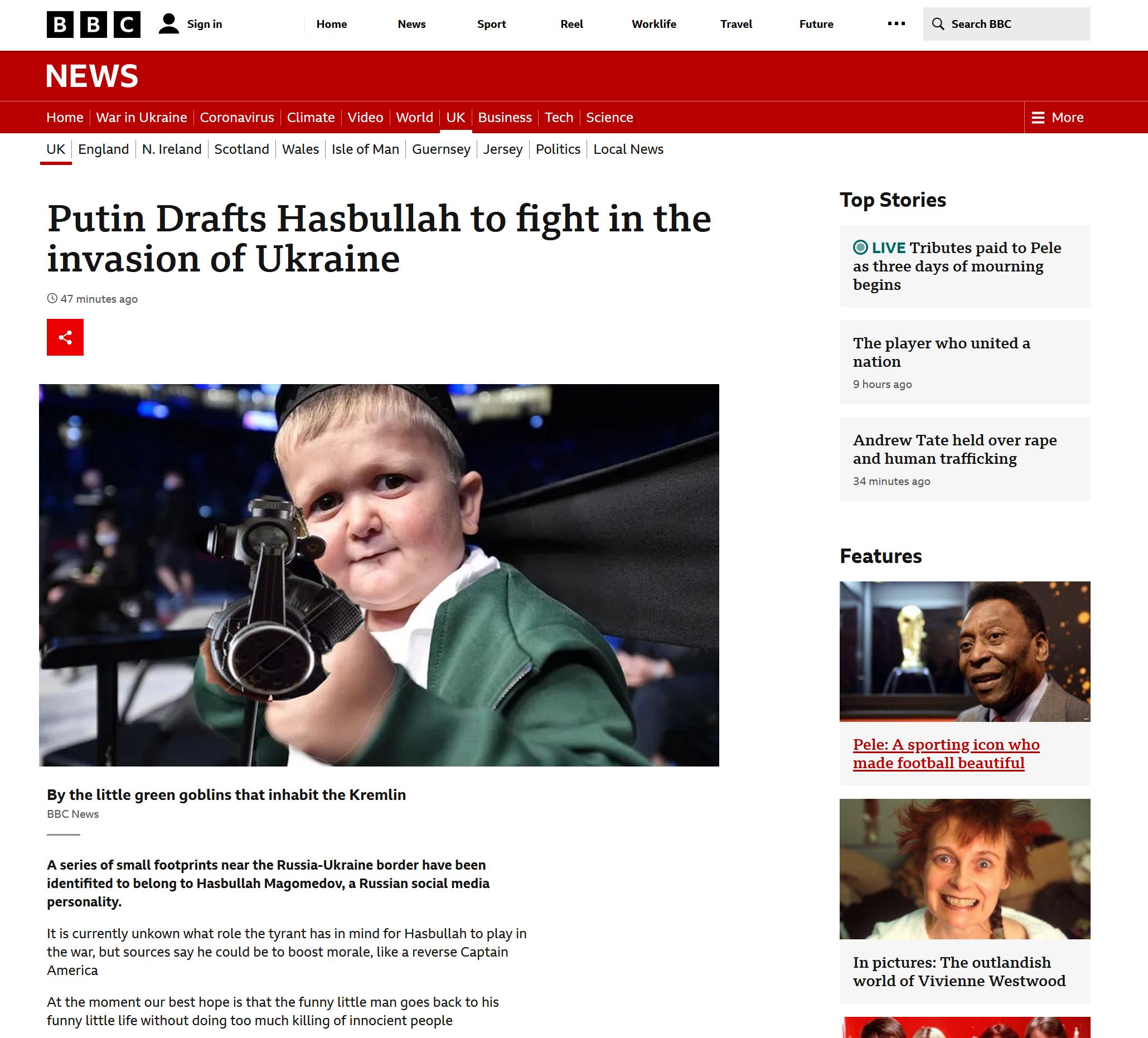} \label{5d}}
\end{tabular}
\caption{Sample no-hate speech images}
\label{nohatespeech}
\end{figure}

Figure \ref{hatespeech} illustrates the samples of text-embedded images corresponding to the label "Hate Speech", while Figure \ref{nohatespeech} illustrates the samples of text-embedded images corresponding to the label "No Hate Speech". An empirical comparison between the actual labels and the predicted labels for the respective image instances quoted in Figures \ref{hatespeech} and \ref{nohatespeech} has been mentioned in Table \ref{empirical}.

With respect to the images corresponding to "Hate Speech", it has been observed for instances \ref{4(b)} and \ref{4(d)} that the labels have been predicted accurately, suggesting the correct prediction capabilities of the model. Similarly, as far as the images corresponding to "No Hate Speech" are concerned, the correct prediction of the labels for instances \ref{5a} and \ref{5b} reemphasize the correct working of the model.

On the contrary, for image instance \ref{4(a)}, the model fails to recognize the implicit attempt of spreading hate speech through visual sarcasm, resulting in a false prediction by the model. Along with, for image instance \ref{4(c)}, hate speech has been embedded in visual form, which was incorrectly detected by the model, leading to another erroneous prediction. This has been precisely due to the lack of ability of the model to decipher the historical context required to detect hate speech in the respective images.

\begin{table}[htbp]
\begin{tabular}{|c|c|c|}
\hline
\textbf{\begin{tabular}[c]{@{}c@{}}Image \\Instance\end{tabular}} & \textbf{Actual Label} & \textbf{Predicted Label} \\ \hline
\ref{4(a)}    & Hate Speech & No Hate Speech \\ \hline
\ref{4(b)}    & Hate Speech & Hate Speech \\ \hline
\ref{4(c)}    & Hate Speech & No Hate Speech \\ \hline
\ref{4(d)}    & Hate Speech & Hate Speech \\ \hline
\ref{5a}    & No Hate Speech & No Hate Speech \\ \hline
\ref{5b}    & No Hate Speech & No Hate Speech \\ \hline
\ref{5c}    & No Hate Speech & Hate Speech \\ \hline
\ref{5d}    & No Hate Speech & Hate Speech \\ \hline
\end{tabular}
\caption{Empirical Evaluation of predictions with respect to sample image instances}
\label{empirical}
\end{table}

As far as the instances in Figure \ref{nohatespeech} are concerned, although instances \ref{5c} and \ref{5d} correspond to the label "No Hate Speech", however, the model faced difficulty in making accurate predictions when confronted with the sarcasm in the image. The prime reason for the model's inaccurate prediction of hate speech in the images is the presence of specific words and phrases that might seem to cause the same at first sight. For instance, \ref{5c} features the words "explosion" and "kills", while \ref{5d} contains the phrase "invasion of Ukraine", which is believed to have been a major cause for this erroneous prediction.

This suggests that the model may require further training or refinement to navigate the nuances of language and accurately identify instances of hate speech more efficiently, even when presented in a lesser straightforward manner.

\section{Conclusion and Future Scope} \label{conclusion}
In this paper, we present our system paper submission for Lexical Squad@Multimodal Hate Speech Event Detection 2023. We aim to classify text-embedded images, indicating whether they contain hate speech or not. The proposed system is an ensemble learning model with fine-tuned InceptionV3, BERT, and XLNet serving as the individual blocks of the proposed model. Given text-embedded images and their respective extracted text through the OCR model, the submitted model classifies each image instance into one of the two labels: "Hate Speech" and "No Hate Speech". The system performs quite well to accomplish the desired task with an accuracy of 75.21\%.

The proposed system can be incorporated for further applications including recommendation systems, personalized content viewing, etc. Along with, it can find usage in further research studies centered on the overlooking field of interest.

In the future, we intend to work on a multitask learning framework to handle social media postings related to other concerns pertaining to sentiment analysis, apart from Hate Speech detection. We also aim to develop models for multi-lingual postings featuring similar scenarios.



\bibliography{custom}
\bibliographystyle{acl_natbib}

\appendix



\end{document}